\documentclass[10pt,twocolumn,letterpaper]{article}

\usepackage[pagenumbers]{cvpr} % To force page numbers, e.g. for an arXiv version

\usepackage{tablefootnote}
\usepackage{comment}
\usepackage{bm}
\usepackage{hhline}
\usepackage{multirow}
\usepackage{color, colortbl}

\definecolor{cvprblue}{rgb}{0.21,0.49,0.74}
\usepackage[pagebackref,breaklinks,colorlinks,allcolors=cvprblue]{hyperref}

\def\paperID{1765} % *** Enter the Paper ID here
\def\confName{CVPR}
\def\confYear{2025}

\title{BioNeRF: Biologically Plausible Neural Radiance Fields for View Synthesis}

\author{Leandro A. Passos, Douglas Rodrigues, Danilo Jodas,\\Kelton A. P. Costa, Jo\~{a}o Paulo Papa\\
Department of Computing, São Paulo State University\\
Av. Eng. Luiz Edmundo Carrijo Coube, 14-01, Bauru, 17033-360, Brazil \\
{\tt\small \{leandro.passos, d.rodrigues, danilo.jodas, kelton.costa, joao.papa\}@unesp.br}
\and
Ahsan Adeel\\
School of Computing, Stirling University\\
Stirling, FK9 4LA, Scotland UK\\
{\tt\small ahsan.adeel@deepci.org}
}

\begin{document}
\maketitle
\begin{abstract}
This paper presents BioNeRF, a biologically plausible architecture that models scenes in a 3D representation and synthesizes new views through radiance fields. Since NeRF relies on the network weights to store the scene's 3-dimensional representation, BioNeRF implements a cognitive-inspired mechanism that fuses inputs from multiple sources into a memory-like structure, improving the storing capacity and extracting more intrinsic and correlated information. BioNeRF also mimics a behavior observed in pyramidal cells concerning contextual information, in which the memory is provided as the context and combined with the inputs of two subsequent neural models, one responsible for producing the volumetric densities and the other the colors used to render the scene. Experimental results show that BioNeRF outperforms state-of-the-art results concerning a quality measure that encodes human perception in two datasets: real-world images and synthetic data.
\end{abstract}    
\section{Introduction}
\label{sec:intro}

Neural Radiance Fields (NeRF)~\cite{mildenhall2021nerf} provide a memory-efficient way to address the problem of rendering new synthetic views based on a set of input images and their respective camera poses. The model implements a Multilayer Perceptron (MLP) network whose weights store a tridimensional scene representation, producing photorealistic quality outputs from new viewpoints while preserving complex features concerning material reflectance and geometry.

Roughly speaking, NeRF comprises a fully connected network that performs regression tasks, i.e., it maps a 5D coordinate sampled from a particular camera viewpoint location relative to the scene, $(x, y, z)$, and their corresponding 2D viewing directions $(\theta, \phi)$, to a view-dependent RGB color, comprising a set of three real-valued numbers, and a single volume density $\Delta$. Such 5D coordinates are transformed using a positional encoding approach to extract higher frequencies, consequently improving the quality of high-resolution representations. Further, the model computes the radiance emitted in such direction using classical volume rendering techniques~\cite{kajiya1984ray}, accumulating those colors and densities into a 2D image. %Figure~\ref{f.nerf} illustrates the concept behind NeRF.

%\begin{figure}[!ht]
%	\centering
%	\includegraphics[width=\textwidth]{figs/nerf.png}
%	\caption{NeRF synthesizes images by sampling 5D coordinates (location and viewing direction) along camera rays (left) and feeding those locations into an MLP to produce color and volume density (right).}
%	\label{f.nerf}
%\end{figure}

One of NeRF's fundamental concepts regards encoding complex real-world 3D scene representation in the parameters of a neural network. Besides, the model restricts the volume density prediction as a function of the camera position, allowing the RGB color to be predicted as a function of both location and viewing direction. Such concepts resemble some more biologically plausible studies inspired by neuroscience discoveries and principles of pyramidal cells~\cite{kording2001supervised}, especially concerning the idea of context to steer the information flow~\cite{adeel2020conscious,kay1998contextually,passos2023multimodal} and integrated memory, responsible for providing additional context based on past experiences~\cite{adeel2022context,passos2023canonical}.% We are unaware of any work that aimed to plug such concepts into the NeRF working mechanism.

This paper proposes BioNeRF, aka Biologically Plausible Neural Radiance Fields, that implements two parallel networks, one to forecast $\Delta$ and the other to predict the RGB color, which interact among themselves providing a context to each other, as well as a  memory mechanism to store intrinsic and correlated information in the learning flow. Such mechanisms are responsible for improving the quality of the generated views and obtaining promising results over two datasets, one composed of synthetic images and the other comprising real images. The main contributions of this work are described as follows:

\begin{itemize}
    \item To propose BioNeRF, a novel and biologically plausible architecture for view synthesis;
    \item To introduce memory and context to view synthesis ambiance;  and
    \item To obtain state-of-the-art results concerning a quality measure that encodes human perception in the context of view synthesis.
\end{itemize}

The remainder of this paper is described as follows. Section~\ref{s.related} presents the related works. Further, Section~\ref{s.proposal} introduces the BioNeRF, while Sections~\ref{s.methodology} and~\ref{s.experiments} describe the methodology and present the experimental results and discussions, respectively. Finally, Section~\ref{s.conclusions} states conclusions and future works.

\section{Related Works}
\label{s.related}

% Before NeRF,  Mildenhall et al.~\cite{mildenhall2019local} proposed the Local Light Field Fusion (LLFF), a method for synthesizing high-quality scene views requiring a significantly smaller number of instances than traditional methods at the time. Meanwhile, Lombardi et al.~\cite{lombardi2019neural} proposed Neural Volumes (NV), a neural volumetric representation method for modeling and rendering dynamic scenes. The technique improved the image resolution without increasing the resolution of the voxel grid. Besides, Sitzmann et al.~\cite{sitzmann2019scene} proposed the Scene Representation Networks (SRNs), which employed unsupervised learning and convolutional neural networks to learn rich representations of 3D scenes without explicit supervision. Further, Liu et al.~\cite{liu2020neural} proposed the Neural Sparse Voxel Fields (NSVF), which rendered dynamic, large-scale scenes using voxel representations and progressive training to obtain high-quality real-time renderings with a reduced time. Finally, in 2021, Mildenhall et al.~\cite{mildenhall2021nerf} proposed the Neural Radiance Fields. This method uses a fully connected neural network to represent scenes as neural radiance fields, synthesizing new views of complex scenes given the camera position and direction and predicting the volumetric densities and emitted color, combined to generate new views using traditional rendering techniques.

Before NeRF,  Mildenhall et al.~\cite{mildenhall2019local} proposed the Local Light Field Fusion (LLFF), a method for synthesizing high-quality scene views requiring a significantly smaller number of instances than traditional methods at the time. Besides, Sitzmann et al.~\cite{sitzmann2019scene} proposed the Scene Representation Networks (SRNs), which employed unsupervised learning and convolutional neural networks to learn rich representations of 3D scenes without explicit supervision.Finally, in 2021, Mildenhall et al.~\cite{mildenhall2021nerf} proposed the Neural Radiance Fields. This method uses a fully connected neural network to represent scenes as neural radiance fields, synthesizing new views of complex scenes given the camera position and direction and predicting the volumetric densities and emitted color, combined to generate new views using traditional rendering techniques.

In the same year, Wang et al.~\cite{wang2021ibrnet} proposed the Image-based Rendering Neural Radiance Field (IBRNet), a learning-based image rendering method whose architecture consists of a multilayer perceptron network and a ray transformer, capable of continuously predicting colors and spatial densities from multiple views and a per-scene fine-tuning procedure. In the meantime, Barron et al.~\cite{barron2021mip} introduced the Multiscale Representation for Anti-Aliasing Neural Radiance Fields (Mip-NeRF). This method extends NeRF to represent the scene at a continuously-valued scale, improving the representation and rendering of three-dimensional scenes and overcoming the limitations of aliasing and blurring. Further, the same authors proposed Mip-NeRF 360~\cite{barron2022mip}, an extension of Mip-NeRF that employs online distillation, non-linear scene parameterization, and distortion-based regularizer to tackle problems related to unbounded scenes present in the prior version.

Chen et al.~\cite{chen2022tensorf} proposed the Tensorial Radiance Fields (TensoRF), a technique that models and reconstructs radiance fields using 4D tensor representation. The work employs the CANDECOMP/PARAFAC decomposition to factorize tensors into compact components and vector-matrix decompositions to relax the component's constraints. Concurrently, Fridovich et al.~\cite{fridovich2022plenoxels} presented the Radiance fields without neural networks (Plenoxels), a system for photorealistic image synthesis that represents a scene as a sparse 3D grid with spherical harmonics. The work of Xu et al.~\cite{xu2022point} introduced the Point-Based Neural Radiance Fields (Point-NeRF), a method for reconstructing and rendering three-dimensional scenes that includes rebuilding a point directly from input images via network inference. Furthermore, Verbin et al.~\cite{verbin2022ref} focused on representing specular reflections in 3D scenes in the Structured View-Dependent Appearance for Neural Radiance Fields (Ref-NeRF). The model employs structured refinements to capture vision-dependent appearance, obtaining state-of-the-art results with photorealistic images generated from new points of view.

Chen et al.~\cite{chen2023local} proposed the Local-to-Global Registration for Bundle-Adjusting Neural Radiance Fields (L2G-NeRF). This method combines local pixel and global frame alignment to achieve high-fidelity reconstruction and addresses large camera pose misalignments. Meanwhile, Kulhanek et al.~\cite{kulhanek2023tetra} presented the Tetra-NeRF, a method for representing neural radiance fields using tetrahedra that combines concepts from 3D geometry processing, triangle-based rendering, point cloud generation, Delaunay triangulation, barycentric interpolation, and modern neural radiance fields. Meanwhile, Yao et al.~\cite{yao2023spiking} proposed the SpikingNeRF, which aligns the radiance rays with the temporal dimension of Spiking Neural Networks, producing an energy-efficient approach associated with biologically plausible inspired concepts. Finally, Selvaratnam et al.~\cite{selvaratnam2024localised} proposed the Localised-NeRF, which projects pixel points onto training images to obtain representation on HSV gradient space
and further extract features to synthesize novel views. The method employs an attention-driven approach to maintain direction consistency and leverage image-based features to predict the diffuse and specular colors, exhibiting competitive performance with prior NeRF-based models.

\section{Biologically Plausible Neural Radiance Fields}
\label{s.proposal}

\begin{figure}[!ht]
	\centering
	\includegraphics[width=\columnwidth]{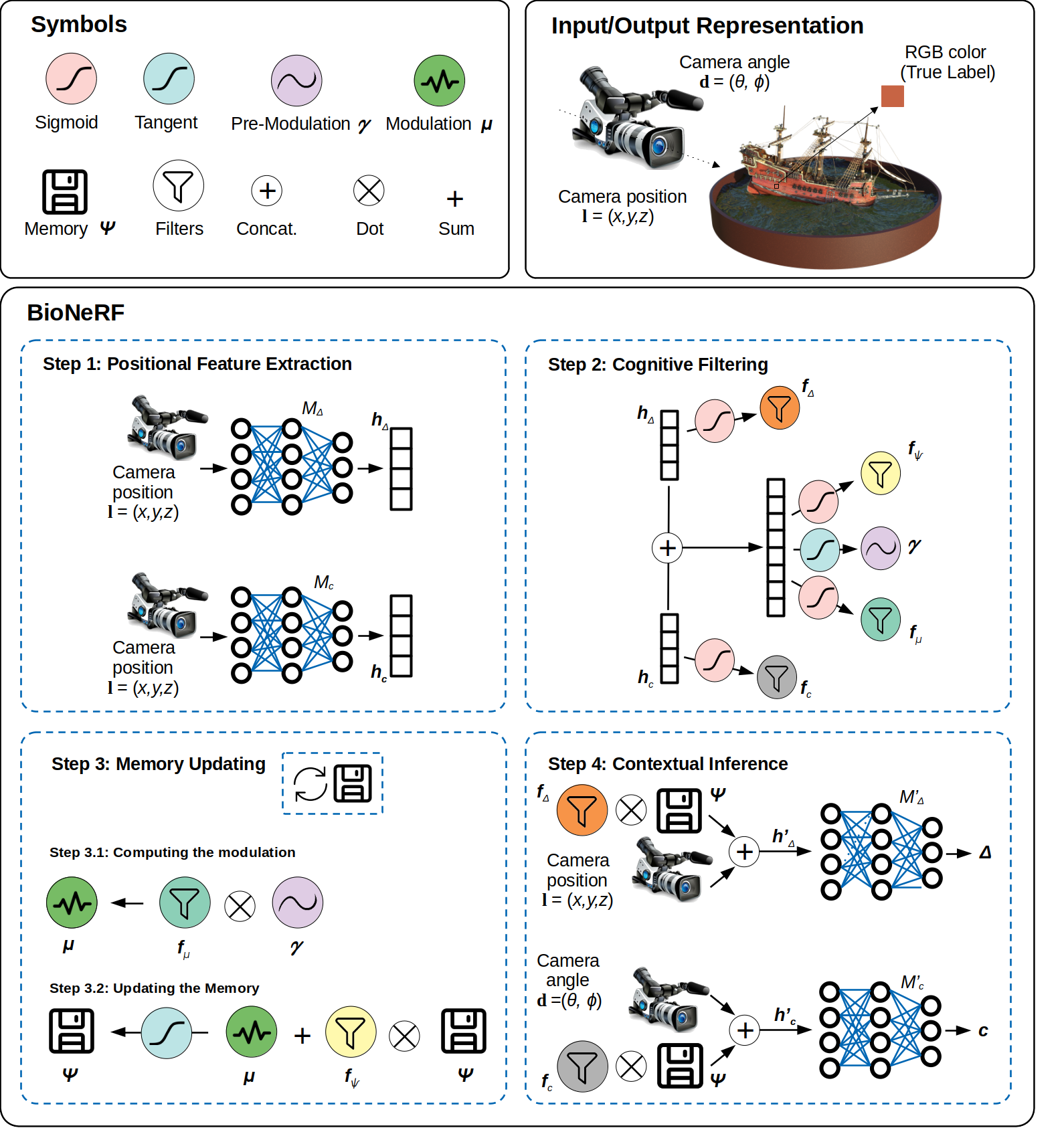}
	\caption{Biologically Plausible Neural Radiance Fields. The top-left frame describes the symbols, while the top-right frames depict the input/output variables. The bottom block illustrates the model's overall pipeline, comprising four steps. Step 1 describes the \textbf{Positional Feature Extraction}, which shall consist of two MLP blocks, namely $M_\Delta$ and $M_c$, responsible for extracting relevant information from the camera input to generate $\bm{h}_\Delta$ and $\bm{h}_c$. Step 2, i.e., \textbf{Cognitive Filtering}, illustrates the filters' generation process, while the \textbf{Memory Updating} in step 3 depicts the memory updating schema. Finally, Step 4, i.e., \textbf{Contextual Inference}, shows how the memory is filtered and concatenated with the camera position to feed $M^\prime_\Delta$ to generate $\Delta$ and combined with the camera angle to feed $M^\prime_c$ to generate $\bm{c}$. Notice that $\Delta$ and $\bm{c}$ are used to render the output compared against the pixel color to compute the BioNeRF's loss.}
	\label{f.bionerf}
\end{figure}

This paper introduces BioNeRF, which combines features extracted from both the camera position and direction into a memory mechanism inspired by cortical circuits~\cite{canonicalLaminar,canonicalCortical}. Further, such memory is employed as a context to leverage the information flow into deeper layers. Figure~\ref{f.bionerf} illustrates the BioNeRF architecture, which spans four main modules: (i) Positional Feature Extraction, (ii) Cognitive Filtering, (iii) Memory Updating, and (iv) Contextual Inference.\newline

\noindent\textbf{Positional Feature Extraction.} The first step consists of feeding two neural models simultaneously, namely $M_{\Delta}$ and $M_c$, such that $\Theta_\Delta = (\bm{W}_\Delta,\bm{b}_\Delta)$ and $\Theta_c = (\bm{W}_c,\bm{b}_c)$ denote their respective set of parameters, i.e., neural weights ($\bm{W}$) and biases ($\bm{b}$). The output of these models, i.e., $\bm{h}_\Delta$ and $\bm{h}_c$, encodes positional information from the input image. Although the input is the same, the neural models do not share weights and follow a different flow in the next steps.\newline

\noindent\textbf{Cognitive Filtering.} This step performs a series of operations from now on called \emph{filters} that work on the embeddings coming from the previous step\footnote{We call those operations \emph{filters}, for they output a value in the interval $[0,1]$ for each feature coming from the embeddings $\bm{h}_\Delta$ and $\bm{h}_c$.}\footnote{Notice such process is inspired in recurrent networks gates' mechanism, like the LSTM~\cite{graves2012long}.}. There are four filters this step derives: (i) density ($\bm{f}_\Delta$), color ($\bm{f}_c$), memory ($\bm{f}_\psi$), and modulation ($\bm{f}_\mu$), computed as follows:

\begin{equation}
	\label{e.f_a}
	%\bm{f}_\Delta = \sigma\left(\bm{W}_\Delta \bm{h}_\Delta+\bm{b_\Delta}\right),
    \bm{f}_\Delta = \sigma\left(\bm{W}_\Delta^f{h}_\Delta+\bm{b}_\Delta^f\right),
\end{equation}

\begin{equation}
	\label{e.f_v}
	%\bm{f}_c = \sigma\left({\bm{W}_c\bm{h}_c+\bm{b_c}}\right),
    \bm{f}_c = \sigma\left(\bm{W}_c^f{h}_c+\bm{b}_c^f\right),
\end{equation}

\begin{equation}
	\label{e.f_m}
	\bm{f}_\psi = \sigma\left({\bm{W}_\psi^f\bm{[h}_\Delta,\bm{h}_c]+\bm{b}_\psi^f}\right),
\end{equation}
and

\begin{equation}
	\label{e.f_w}
	\bm{f}_\mu = \sigma\left({\bm{W}_\mu^f[\bm{h}_\Delta,\bm{h}_c]+\bm{b}_\mu^f}\right),
\end{equation}
where $\bm{W}_\Delta^f$, $\bm{W}_c^f$, $\bm{W}_\psi^f$ and $\bm{W}_\mu^f$ correspond to the weight matrices for the density, color, memory, and modulation filters, respectively. Additionally, $\bm{b_\Delta^f}$, $\bm{b_c^f}$, $\bm{b_\psi^f}$ and $\bm{b_\mu^f}$ stand for their respective biases. Moreover, $[\bm{h}_\Delta,\bm{h}_c]$ represents the concatenation of embeddings $\bm{h}_\Delta$ and $\bm{h}_c$, while $\sigma$ denotes a sigmoid function. The pre-modulation $\bm{\gamma}$ is computed as follows:

%where $\bm{W}_\Delta$, $\bm{W}_c$, $\bm{W}_\psi$, and $\bm{W}_\mu$ correspond to the weight matrices for the density, color, memory, and modulation filters, respectively. Additionally, $\bm{b_\Delta}$, $\bm{b_f}$, $\bm{b_\psi}$, and $\bm{b_\mu}$ stand for their respective biases. Moreover, $[\bm{h}_\Delta,\bm{h}_c]$ represents the concatenation of embeddings $\bm{h}_\Delta$ and $\bm{h}_c$, while $\sigma$ denotes a Sigmoid function. The pre-modulation $\bm{\gamma}$ is computed as follows, where :

\begin{equation}
	\label{e.gamma}
	\bm{\gamma} = tanh\left({\bm{W}_{\gamma}[\bm{h}_\Delta,\bm{h}_c]+\bm{b}_{\gamma}}\right),
\end{equation}
where $tanh(\cdot)$ is the hyperbolic tangent function, while $\bm{W}_{\gamma}$ and $\bm{b}_{\gamma}$ are the pre-modulation weight matrix and bias, respectively.\newline

\noindent\textbf{Memory Updating.} Updating the memory requires the implementation of a mechanism capable of obliterating trivial information, which is performed using the memory filter $\bm{f}_\psi$ (Step 3.1 in Figure~\ref{f.bionerf}). Fist, one needs to compute the modulation $\bm{\mu}$, where $\otimes$ represents the dot product:

\begin{equation}
	\label{e.omega}
	\bm{\mu} = \bm{f}_\mu \otimes \bm{\gamma}.
\end{equation}

New experiences are introduced in the memory $\bm{\Psi}$ through the modulating variable $\bm{\mu}$ using a $\textit{tanh}$ function (Step 3.2 in Figure~\ref{f.bionerf}):

% \begin{equation}
% 	\label{e.mu_n}
% 	\Psi = \mu + \left(f_\psi\otimes \Psi\right),
% \end{equation}

\begin{equation}
	\label{e.psi}
	\bm{\Psi} = tanh\left({\bm{W}_{\Psi}\left( \bm{\mu} + \left(\bm{f}_\psi\otimes \Psi\right) \right)+\bm{b}_{\Psi}}\right),
\end{equation}
where $\bm{W}_{\Psi}$ and $\bm{b}_{\Psi}$ are the memory weight matrix and bias, respectively. \newline

\noindent\textbf{Contextual Inference.} This step is responsible for adding contextual information to BioNeRF. We generate two new embeddings $\bm{h}^{\prime}_\Delta$ and $\bm{h}^{\prime}_c$ based on filters $\bm{f}_\Delta$ and $\bm{f}_c$, respectively (Step 4 in Figure~\ref{f.bionerf}), which further feed two neural models, i.e., $M^{\prime}_\Delta=(\bm{W}^{\prime}_\Delta,\bm{b}^{\prime}_\Delta)$ and $M^{\prime}_c=(\bm{W}^{\prime}_c,\bm{W}^{\prime}_c)$, accordingly:

%In the sequence, $M_\Delta$ and $M_c$ outputs, namely $\bm{h}_\Delta$ and $\bm{h}_c$, respectively, are updated to accommodate information provided from both the memory and the context, i.e., the contribution of $\bm{h}_\Delta$ to foster $M^{\prime}_c$ and $\bm{h}_c$ to leverage $M^{\prime}_\Delta$ convergence. The procedure is computed as follows:

\begin{equation}
	\label{e.h_a_line}
	\bm{h}_\Delta' = [\bm{\Psi} \otimes \bm{f}_\Delta,\mathbf{l}],
\end{equation}
and
\begin{equation}
	\label{e.h_v_line}
	\bm{h}_c' = [\bm{\Psi} \otimes \bm{f}_c,\mathbf{d}].
\end{equation}

Subsequently, $M^{\prime}_\Delta$ outputs the volume density $\Delta$, while color information $\bm{c}$ is predicted by $M^{\prime}_c$, ending up in the final predicted pixel information $(\Delta,\bm{c})$, further used to compute the loss function.\newline

\noindent\textbf{Loss Function.} Let $r:\Re\times\Re^3\rightarrow\Re^3$ be a volume rendering technique~\cite{kajiya1984ray} that computes the pixel color given the volume density and the color. The BioNeRF loss function is defined as follows:

\begin{equation}
	\label{e.loss}
	{\cal L} = MSE(r(\Delta, \bm{c}), \bm{g}),
\end{equation}
where $MSE(\cdot)$ is the mean squared error function and $\bm{g}$ corresponds to the ground truth pixel color. 

The error is then back-propagated to the model's previous layers/steps to update its set of weights ($\bm{W}=\{\bm{W}_\Delta,\bm{W}_c,\bm{W}_\Delta^f,\bm{W}_c^f,\bm{W}_\psi^f$,$\bm{W}_\mu^f,\bm{W}_{\gamma},\bm{W}_{\Psi},\bm{W}^{\prime}_\Delta,\bm{W}^{\prime}_c\}$) and biases $\bm{b}=\{\bm{b}_\Delta,\bm{b}_c,\bm{b}_\Delta^f,\bm{b}_c^f,\bm{b}_\psi^f$,$\bm{b}_\mu^f,\bm{b}_{\gamma},\bm{b}_{\Psi},\bm{b}^{\prime}_\Delta,\bm{b}^{\prime}_c\}$.

\section{Methodology}
\label{s.methodology}

This section provides information regarding the datasets employed in this work and the configuration adopted in the experimental setup.

\subsection{Datasets}

We conducted experiments over two well-known datasets concerning view synthesis, i.e., Blender~\cite{mildenhall2021nerf} and LLFF~\cite{mildenhall2019local}:

\subsubsection{Blender}

Also known as NeRF Synthetic, the Blender comprises eight object scenes with intricate geometry and realistic non-Lambertian materials. Six of these objects are rendered from viewpoints tested on the upper hemisphere, while the remaining two come from viewpoints sampled on a complete sphere. 

\subsubsection{Local Light Field Fusion} 

This paper considers $8$ scenes from the LLFF Real dataset, which comprises $24$ scenes captured from handheld cellphones with $20-30$ images each. The authors used a COLMAP structure from motion~\cite{schonberger2016structure} implementation to compute the poses.

\subsection{Experimental Setup}

The experiments conducted in this work aim to evaluate the behavior of BioNeRF in the context of scene-view synthesis against state-of-the-art methods. In BioNeRF, the camera 3D coordinates feed two neural models, i.e., blocks of dense layers with ReLU activations, each comprising $3$ layers with $h=256$ hidden neurons. Notice that the number of layers and hidden neurons were empirically selected based on values adopted in the original NeRF model. Using separate blocks resembles the biological brain's efficiency in fusing information from multiple sources. Further, the mechanism described in Section~\ref{s.proposal} is performed to generate the filters and update the memory $\Psi\in\mathbb{R}^{z\times h}$, where $z=8,192$ is the number of directional rays processed in parallel.

In the sequence, the memory is used as a context to the consecutive blocks, being concatenated with the camera 3D coordinates to feed the second neural model $M^\prime_\Delta$, which comprises two dense layers with $256$ neurons and an output layer with a single neuron to predict the volume density $\Delta$. Similarly, the memory is concatenated with the 2D viewing direction $(\theta, \phi)$ to feed the second model $M^\prime_c$, which is composed of a dense layer with $128$ neurons and an output layer with three neurons to predict the directional emitted color. The network parameters were selected empirically based on the methodology employed in~\cite{mildenhall2021nerf}, consisting of the Adam optimizer with a learning rate of $5e-4$ during $400k$ updates. 

Additionally, three quality measures were considered to evaluate the models: (i) the Peak Signal Noise Ratio (PSNR), (ii) the Structural Similarity Index Measure (SSIM), and (iii) the Learned Perceptual Image Patch Similarity (LPIPS). PSNR defines the ratio between the maximum power of a signal and the noise that affects the signal representation and is used to quantify a signal's reconstruction. SSIM predicts the perceived quality of digital images, considering its degradation as a change in the structural information. Lastly, LPIPS computes the similarity between the activations of two image patches for some predefined network, presenting itself as an adequate metric to match human perception. 

The experiments were conducted on an Ubuntu 18 system running on a 2x Intel\textsuperscript{\textregistered} Xeon Bronze 3104 processor with $62$GB of memory and an NVIDIA\textsuperscript{\textregistered} Tesla T4 GPU. The code was implemented using Python and the PyTorch framework and is available on GitHub\footnote{Ommited due to anonymous review.}.%\footnote{\url{https://github.com/Leandropassosjr/BioNeRF}.}.

\section{Experiments}
\label{s.experiments}

This section evaluates BioNeRF concerning view synthesis against state-of-the-art procedures over two datasets comprising real and synthetic images.

\subsection{General Evaluation}
\label{s.general}

Table~\ref{t.experiments_general} provides the general quantitative results for view synthesis, comparing BioNeRF against state-of-the-art methods considering both synthetic and real image datasets\footnote{Values marked with $-$ describe results not provided by the authors.}. The analysis is conducted based on three image reconstruction metrics: PSNR and SSIM, whose objective is to be maximized, and LPIPS, whose lower values denote better results. Notice that bolded numbers represent the most accurate values, while the colors red, orange, and yellow denote the first, second, and third best results, respectively.

\begin{table*}[!htb]
\caption{General results considering both synthetic and real image datasets.}
\begin{center}
\renewcommand{\arraystretch}{1.5}
\setlength{\tabcolsep}{6pt}
\resizebox{0.8\textwidth}{!}{
\begin{tabular}{|clccccccccc|}
\hhline{|-|-|-|-|-|-|-|-|-|-|-|}
&&&\multicolumn{3}{c}{\textbf{Blender\cite{mildenhall2021nerf}}}&&\multicolumn{3}{c}{\textbf{LLFF (Real)\cite{mildenhall2019local}}}&\\\cline{4-6}\cline{8-10} 
&{\textbf{Method}} && \textbf{PSNR }$\uparrow$ & \textbf{SSIM }$\uparrow$ & \textbf{LPIPS }$\downarrow$ & &\textbf{PSNR }$\uparrow$ & \textbf{SSIM }$\uparrow$ & \textbf{LPIPS }$\downarrow$ &\\ \cline{2-10} 
&SRN~\cite{sitzmann2019scene} (NeurIPS'19)  && $22.26$ & $0.846$ & $0.170$ && $22.84$ & $0.668$ & $0.378$ &\\
% &NV~\cite{lombardi2019neural} (ACM ToG 19) && $26.05$ & $0.893$ & $0.160$ && $-$ & $-$ & $-$ & \\
&LLFF~\cite{mildenhall2019local} (ACM ToG 2019)&& $24.88$ & $0.911$ & $0.114$ && $24.13$ & $0.798$ & $0.212$ &\\

% &NSVF~\cite{liu2020neural} (NeurIPS'20)  && $31.75$ & $0.964$ & $0.047 $ && $-$ & $-$ & $-$  &\\

&NeRF~\cite{mildenhall2021nerf} (ACM 21) && $31.01$ & $0.947$ & $0.081$ && $26.50$ & $0.811$ & $0.250$ &\\

&IBRNet~\cite{wang2021ibrnet} (CVPR'21)  && $28.14 $ & $0.942$ & $0.072$ && \cellcolor{orange!25}$26.73$ & \cellcolor{orange!25}$0.851$ & $0.175$  &\\

&Mip-NeRF~\cite{barron2021mip} (ICCV'21)  && $33.09$ & $0.961$ & $0.043 $ && $-$ & $-$ & $-$  &\\

&Mip-NeRF 360~\cite{barron2022mip} (CVPR'22)  && $30.34$ & $0.951$ & $0.060 $ && $26.59$ & \cellcolor{yellow!25}$0.846$ & \cellcolor{yellow!25}$0.168$  &\\

%&R2-L\cite{wang2022r2l} (2022) && $31.87$ & $0.950$ & $0.034$ && $27.79$ & $0.972$ & $0.096$ &\\
&TensoRF~\cite{chen2022tensorf} (ECCV'22) && $33.14$ & $0.963$ & \cellcolor{orange!25}$0.027$ && \cellcolor{orange!25}$26,73$ & $0.839$ & \cellcolor{orange!25}$0.124$ &\\
&Plenoxels~\cite{fridovich2022plenoxels} (CVPR'22) && $31.71$ & $0.958$ & $0.049$ && $26,29$ & $0.839$ & $0.210$ &\\
&Point-NeRF~\cite{xu2022point} (CVPR'22)  && \cellcolor{orange!25}$33.31$ & \cellcolor{orange!25}$0.978$ & \cellcolor{orange!25}$0.027$ && $-$ & $-$ & $-$  &\\

&Ref-NeRF~\cite{verbin2022ref} (CVPR'22)  && $\cellcolor{red!25}\bm{33.99} $ & $0.966$ & $0.038 $ && $-$ & $-$ & $-$  &\\

&L2G-NeRF~\cite{chen2023local} (CVPR'23) && $28.62$ & $0.930$ & $0.070$ && $24.54$ & $0.750$ & $0.200$ &\\
&Tetra-NeRF~\cite{kulhanek2023tetra} (ICCV'23) && $32.53 $ & $\cellcolor{red!25}\bm{0.982}$ & $0.041$ && $-$ & $-$ & $-$  &\\
%&NeRF-PR~\cite{heo2023robust} (2023) && $29.86$ & $0.943$ & $0.056$ && $24.79$ & $0.772$ & $0.197$ &\\ \cline{2-10} 

&SpikingNeRF~\cite{yao2023spiking} (arXiv'23) && $32.45$ & $0.956$ & $-$ && $-$ & $-$ & $-$  &\\

&Localised-NeRF:~\cite{selvaratnam2024localised} (CVPR'24) && \cellcolor{yellow!25}$33.25$ & \cellcolor{yellow!25}$0.969$ & $-$ && $-$ & $-$ & $-$  &\\\cline{2-10}

&BioNeRF (Ours)  && $31.45$ & $0.953$ & $\cellcolor{red!25}\bm{0.026}$ && \cellcolor{red!25}$\bm{27.01}$ & $\cellcolor{red!25}\bm{0.861}$ & \cellcolor{red!25}$\bm{0.068}$ &\\
\hhline{|-|-|-|-|-|-|-|-|-|-|-|}
\end{tabular}}
\label{t.experiments_general}
\end{center}
\end{table*}

The results confirm the effectiveness of BioNeRF since it could obtain the best outcomes for all three metrics considering the LLFF dataset, which comprises $8$ scenes from a real environment. BioNeRF also obtained the lowest LPIPS value over the synthetic dataset, outperforming all state-of-the-art results in this context. Such results are highly positive since LPIPS correspond to human perceptual judgments far better than the commonly used metrics like PSNR and SSIM~\cite{zhang2018unreasonable}.

The win-win results of BioNeRF are primarily due to its memory and context components. Since NeRF relies on the network weights to store the 3D representation of a scene, it makes sense to introduce a memory mechanism that is updated by keeping more relevant information and discarding what is unessential. Further, using such memory as a context forces the model to associate correlated features and produce coherent view-dependent output colors and volume densities, thus improving its performance.

\subsection{Quantitative evaluation}
\label{ss.quantitative}

% \subsubsection{Blender dataset:}
% \label{sss.qualitativeBlender}

Table~\ref{t.results_Synthetic} presents results concerning the Blender dataset. Regarding the PSNR measure, one can observe that Ref-NeRF obtained the most accurate average results, achieving higher PSNR values over half of the scenes, while Point-NeRF stood out in two of them, and TensoRF and Tetra-NeRF did well in Lego and Ship, respectively. Concerning the SSIM metric, even though Point-NeRF obtained the best results on six out of eight scenes, Tetra-NeRF achieved the best average accuracy overall due to an excellent performance over the Ship scene.

\begin{table*}[!htb]
\caption{Quantitative results considering each scene from Blender dataset.}
\begin{center}
\renewcommand{\arraystretch}{1.5}
\setlength{\tabcolsep}{6pt}
\resizebox{0.8\textwidth}{!}{
\begin{tabular}{|clcccccccccccc|}
\hhline{|-|-|-|-|-|-|-|-|-|-|-|-|-|-|}
&&&\multicolumn{10}{c}{\textbf{PSNR }$\uparrow$ }&\\
&{\textbf{Method}} && Avg. &|&Chair &Drums& Ficus& Hotdog& Lego& Materials& Mic &Ship &\\ \cline{2-13} 
%&SRN~\cite{sitzmann2019scene} (2019)  && $22.26$ &|& $26.96$ & $17.18$ & $20.73$ & $26.81$ & $20.85$ & $ 18.09$ & $26.85$ &  $20.60$ &\\
%&NV~\cite{lombardi2019neural} (2019) && $0$ && $0$ & $0$ & $0$ & $0$ & $0$ & $0$ & $0$ & $0$ & \\
%&LLFF~\cite{mildenhall2019local} (2019 ) && $0$ && $0$ & $0$ & $0$ & $0$ & $0$ & $0$ & $0$ & $0$ & \\

%&NSVF~\cite{liu2020neural} (2020)   && $31.75$ &|& $33.19$ & $25.18$ & $31.23$ & $37.14$ & $32.29$ & $32.68$ & $34.27$ & $27.93$ & \\

&NeRF~\cite{mildenhall2021nerf} (ACM 21)  && $31.01$ &|& $33.00$ & $25.01$ & $30.13$ & $36.18$ & $32.54$ & $29.62$ & $32.91$ & $28.65$ & \\

%&IBRNet~\cite{wang2021ibrnet} (CVPR'21)   && $0$ && $0$ & $0$ & $0$ & $0$ & $0$ & $0$ & $0$ & $0$ & \\

%&Mip-NeRF~\cite{barron2021mip} (ICCV'21)  && $0$ && $0$ & $0$ & $0$ & $0$ & $0$ & $0$ & $0$ & $0$ & \\

&TensoRF~\cite{chen2022tensorf} (ECCV'22)  &&\cellcolor{yellow!25} $33.14$ &|& \cellcolor{orange!25}$35.76$ & $ \cellcolor{orange!25}26.01$ & $\cellcolor{orange!25} 33.99$ & \cellcolor{orange!25}$37.41$ & $\cellcolor{red!25}\bm{36.46}$ & \cellcolor{orange!25}$30.12$ & $34.61$ & \cellcolor{yellow!25}$30.77$ & \\
&Plenoxels~\cite{fridovich2022plenoxels} (CVPR'22)  && $31.71$ &|& $33.98$ & $25.35$ & $31.83$ & $36.43$ & $34.10$ & $29.14$ & $33.26$ & $29.62$ & \\
&Point-NeRF~\cite{xu2022point} (CVPR'22)  && \cellcolor{orange!25}$33.31$ &|& \cellcolor{yellow!25}$35.40$ & $\cellcolor{red!25}\bm{26.06}$ & $\cellcolor{red!25}\bm{36.13}$ & \cellcolor{yellow!25}$37.30$ & \cellcolor{yellow!25}$35.04$ & $29.61$ & \cellcolor{orange!25}$35.95$ & \cellcolor{orange!25}$30.97$ & \\

&Ref-NeRF~\cite{verbin2022ref} (CVPR'22)   &&\cellcolor{red!25}$\bm{33.99}$ &|& $ \cellcolor{red!25}\bm{35.83}$ &\cellcolor{yellow!25} $25.79$ & \cellcolor{yellow!25}$33.91 $ & $\cellcolor{red!25}\bm{37.72}$ & \cellcolor{orange!25}$36.25 $ & $\cellcolor{red!25}\bm{35.41} $ & $\cellcolor{red!25}\bm{36.76}$ & $30.28$ & \\

&Mip-NeRF 360~\cite{barron2022mip} (CVPR'22)  && $30.34$ &|& $34.01$ & $24.36$ & $26.66$ & $36.44$ & $33.20 $ & $27.91$ & $31.50$ & $28.66$ & \\

&L2G-NeRF~\cite{chen2023local} (CVPR'23)  && $28.62$ &|& $30.99$ & $23.75 $ & $26.11 $ & $34.56 $ & $27.71 $ & $27.60 $ & $30.91$ & $27.31 $ & \\
&Tetra-NeRF~\cite{kulhanek2023tetra} (ICCV'23)  && $32.53$ &|& $35.05$ & $25.01$ & $33.31$ & $36.16$ & $34.75$ & $29.30$ & \cellcolor{yellow!25}$35.49$ & $\cellcolor{red!25}\bm{31.13}$ & \\\cline{2-13} 
%&NeRF-PR~\cite{heo2023robust} (2023)  && $0$ && $0$ & $0$ & $0$ & $0$ & $0$ & $0$ & $0$ & $0$ & \\ \cline{2-10} 

&BioNeRF (Ours)  && $31.45 $ &|& $34.63$ & $25.66$ & $29.56$ & $37.23$ & $31.82$ & \cellcolor{yellow!25}$29.74$ & $33.38$ & $29.57$ & \\
% &BioNeRF (Ours)  && $30.85 $ &|& $32.79$ & $24.93$ & $30.09$ & $36.12$ & $32.53$ & $29.54$ & $32.41$ & $28.39$ & \\
\hhline{|-|-|-|-|-|-|-|-|-|-|-|-|-|-|}
\end{tabular}}

\vspace{0.1cm}

\resizebox{0.8\textwidth}{!}{
\begin{tabular}{|clcccccccccccc|}
\hhline{|-|-|-|-|-|-|-|-|-|-|-|-|-|-|}
&&&\multicolumn{10}{c}{\textbf{SSIM }$\uparrow$ }&\\
&{\textbf{Method}} && Avg. &|&Chair &Drums& Ficus& Hotdog& Lego& Materials& Mic &Ship &\\ \cline{2-13} 
&NeRF~\cite{mildenhall2021nerf} (ACM 21)  && $0.947$ &|& $0.967$ & $0.925$ & $0.964$ & $0.974$ & $0.961$ & $0.949$ & $0.980$ & $0.856$ &\\
&TensoRF~\cite{chen2022tensorf} (ECCV'22)  && $0.963 $ &|& \cellcolor{yellow!25}$0.985$ & $0.937$ &\cellcolor{yellow!25} $0.982$ & $0.982$ & \cellcolor{yellow!25}$0.983 $ & $0.952$ & $0.988$ & $0.895$ & \\
&Plenoxels~\cite{fridovich2022plenoxels} (CVPR'22)  && $0.958$ &|& $0.977$ & $0.933$ & $0.890$ & \cellcolor{yellow!25}$0.985$ & $0.976 $ &\cellcolor{orange!25}$0.975$ & $0.980 $ & $\cellcolor{orange!25}0.949$ & \\\
&Point-NeRF~\cite{xu2022point} (CVPR'22) && \cellcolor{orange!25}$0.978$ &|& $\cellcolor{red!25}\bm{0.991}$ & $\cellcolor{red!25}\bm{0.954} $ & $\cellcolor{red!25}\bm{0.993}$ & $\cellcolor{red!25}\bm{0.991} $ & $\cellcolor{red!25}\bm{0.988} $ & \cellcolor{yellow!25}$0.971 $ & $\cellcolor{red!25}\bm{0.994} $ & \cellcolor{yellow!25}$0.942 $ & \\
&Ref-NeRF~\cite{verbin2022ref} (CVPR'22)  && \cellcolor{yellow!25}$0.966$ &|& $0.984$ & \cellcolor{yellow!25}$0.937$ & $0.983$ & $0.984$ & $0.981 $ & $\cellcolor{red!25}\bm{0.983}$ & \cellcolor{orange!25}$0.992 $ & $0.880$ & \\

&Mip-NeRF 360~\cite{barron2022mip} (CVPR'22)  && $0.951$ &|& $0.977$ & $0.923$ & $0.952$ & $0.979$ & $0.975 $ & $0.944$ & $0.984 $ & $0.875$ & \\

&L2G-NeRF~\cite{chen2023local} (CVPR'23)  && $0.930$ &|& $0.950$ & $0.900$ & $ 0.930$ & $0.970$ & $0.910$ & $0.930$ & $0.970$ & $0.850$ & \\
&Tetra-NeRF~\cite{kulhanek2023tetra} (ICCV'23)  && $\cellcolor{red!25}\bm{0.982}$ &|& \cellcolor{orange!25}$0.990 $ & $\cellcolor{orange!25}0.947$ & \cellcolor{orange!25}$0.989 $ & $\cellcolor{orange!25}0.989$ & $\cellcolor{orange!25}0.987$ & $0.968 $ & \cellcolor{yellow!25}$0.993 $ & $\cellcolor{red!25}\bm{0.994}$ & \\\cline{2-13} 
%&NeRF-PR~\cite{heo2023robust} (2023)  && $0$ && $0$ & $0$ & $0$ & $0$ & $0$ & $0$ & $0$ & $0$ & \\ \cline{2-10} 

&BioNeRF (Ours)  && $0.953$ &|& $0.977$ & $0.927$ & $0.965$ & $0.980$ & $0.963$ & $0.957$ & $0.978$ & $0.874$ & \\
% &BioNeRF (Ours)  && $0.945$ &|& $0.965$ & $0.921$ & $0.962$ & $0.974$ & $0.960$ & $0.948$ & $0.978$ & $0.856$ & \\
\hhline{|-|-|-|-|-|-|-|-|-|-|-|-|-|-|}
\end{tabular}}

\vspace{0.1cm}

\resizebox{0.8\textwidth}{!}{
\begin{tabular}{|clcccccccccccc|}
\hhline{|-|-|-|-|-|-|-|-|-|-|-|-|-|-|}
&&&\multicolumn{10}{c}{\textbf{LPIPS }$\downarrow$ }&\\
&{\textbf{Method}} && Avg. &|&Chair &Drums& Ficus& Hotdog& Lego& Materials& Mic &Ship &\\ \cline{2-13} 
&NeRF~\cite{mildenhall2021nerf} (ACM 21)  && $0.081$ &|& $0.046$ & $0.091$ & $0.044$ & $0.121$ & $0.050$ & $0.063$ & $0.028$ & $0.206 $ &\\
&TensoRF~\cite{chen2022tensorf} (ECCV'22)  && \cellcolor{orange!25}$0.027$ &|& $\cellcolor{red!25}\bm{0.010}$ & $\cellcolor{orange!25}0.051$ & $\cellcolor{red!25}\bm{0.012}$ & $\cellcolor{orange!25}0.013$ & $\cellcolor{red!25}\bm{0.007}$ & \cellcolor{yellow!25}$0.026$ & $\cellcolor{orange!25}0.009$ & $\cellcolor{orange!25}0.085$ & \\
&Plenoxels~\cite{fridovich2022plenoxels} (CVPR'22)  && $0.049$ &|& $0.031 $ & $0.067$ & $0.026 $ & $0.037$ & $0.028$ & $0.057$ & $0.015$ & $0.134$ & \\
&Point-NeRF~\cite{xu2022point} (CVPR'22) && $0.049$ &|& $0.023$ & $0.078$ & $0.022$ & $0.037$ & $0.024$ & $0.072 $ & $0.014$ & $0.124$ & \\
&Ref-NeRF~\cite{verbin2022ref} (CVPR'22)  && \cellcolor{yellow!25}$0.038 $ &|& $0.017 $ & $\cellcolor{yellow!25}0.059$ & $\cellcolor{yellow!25}0.019$ & \cellcolor{yellow!25}$0.022 $ & $\cellcolor{yellow!25}0.018 $ & $\cellcolor{orange!25}0.022$ & $\cellcolor{red!25}\bm{0.007} $ & $0.139$ & \\

&Mip-NeRF 360~\cite{barron2022mip} (CVPR'22)  && $0.060$ &|& $0.032$ & $0.083$ & $0.048$ & $0.039$ & $0.028 $ & $0.067$ & $0.021 $ & $0.164$ & \\

&L2G-NeRF~\cite{chen2023local} (CVPR'23)  && $0.070$ &|& $0.050$ & $0.100$ & $0.060$ & $0.030$ & $0.060$ & $0.060$ & $0.050$ & $0.130$ & \\
&Tetra-NeRF~\cite{kulhanek2023tetra} (ICCV'23)  && $0.041$ &|& $\cellcolor{yellow!25}0.016 $ & $0.073$ & $0.023$ & $0.027$ & $0.022 $ & $0.056$  &\cellcolor{yellow!25} $0.011 $ & $\cellcolor{yellow!25}$0.103$ $ & \\\cline{2-13} 
%&NeRF-PR~\cite{heo2023robust} (2023)  && $0$ &|& $0$ & $0$ & $0$ & $0$ & $0$ & $0$ & $0$ & $0$ & \\ \cline{2-10} 

&BioNeRF (Ours)  && \cellcolor{red!25}$\bm{0.026}$ &|& \cellcolor{orange!25}$0.011$ & \cellcolor{red!25}$\bm{0.047}$ & \cellcolor{orange!25}$0.017$ & \cellcolor{red!25}$\bm{0.010}$ & \cellcolor{orange!25}$0.016$ & \cellcolor{red!25}$\bm{0.018}$ & $0.018$ & \cellcolor{red!25}$\bm{0.068}$ & \\
% &BioNeRF (Ours)  &&\cellcolor{orange!25} $0.038$ &|& $0.027$ & $\cellcolor{yellow!25}0.063$ & $\cellcolor{yellow!25}0.020$ & $\cellcolor{orange!25}0.020$ & \cellcolor{yellow!25}$0.019$ & $\cellcolor{orange!25}0.025$ & $0.018$ & $\cellcolor{yellow!25}0.113$ & \\
\hhline{|-|-|-|-|-|-|-|-|-|-|-|-|-|-|}
\end{tabular}}
\label{t.results_Synthetic}
\end{center}
\end{table*}

Apart from such results, BioNeRF asserted the average best results considering the LPIPS measure, obtaining the minimum value over four out of eight scenes, i.e., Drums, Hotdog, Materials, and Ship. Further, it virtually broke even with the best results considering the Chair scene, reaching an LPIPS of $0.011$ against $0.010$ from TensoRF. Such results are very positive since LPIPS is the measure that best corresponds to human perceptual judgments, thus implying better visual appearance.

% \subsubsection{LLFF Real dataset:}
% \label{sss.qualitativeLLFF}

BioNeRF obtained paramount results concerning the LLFF dataset, surpassing the state of the art for all measures, as presented in Table~\ref{t.results_real}. The method received the highest average PSNR value, reaching the best results in five out of eight scenes, while TensoRF performed better in Flower and Horn's scenes, and NeRF obtained the best results in the Room scenes.

Concerning the SSIM and LPIPS measures, BioNeRF achieved the best results overall, outperforming other techniques in every scene. Such results confirm the robustness of BioNeRF and the relevance of the context to steer the information flow and extract coherent features that lead to more adequate outputs, as well as the memory mechanism responsible for filtering the relevant knowledge and focusing on proper representations.

\begin{table*}[!htb]
\caption{Quantitative results considering each scene from LLFF Real dataset.}
\begin{center}
\renewcommand{\arraystretch}{1.5}
\setlength{\tabcolsep}{6pt}
\resizebox{0.8\textwidth}{!}{
\begin{tabular}{|clcccccccccccc|}
\hhline{|-|-|-|-|-|-|-|-|-|-|-|-|-|-|}
&&&\multicolumn{10}{c}{\textbf{PSNR }$\uparrow$ }&\\
&{\textbf{Method}} && Avg. &|&Fern&Flower&Fortress&Horns &Leaves& Orchids& Room &   T-Rex &\\ \cline{2-13} 
&NeRF~\cite{mildenhall2021nerf} (ACM 21)   && $26.50$ &|& $25.17$ & $27.40$ & $31.16$ & $27.45$ & $20.92$ & \cellcolor{orange!25}$20.36$ & \cellcolor{orange!25}$32.70 $ & $26.80$ &  \\
&TensoRF~\cite{chen2022tensorf} (ECCV'22)   && $26.73 $ &|& \cellcolor{yellow!25}$25.27 $ & \cellcolor{red!25}$\bm{28.60}$ & \cellcolor{orange!25}$31.36 $ & \cellcolor{orange!25}$28.14$ & \cellcolor{yellow!25}$21.30$ & $19.87$ & \cellcolor{yellow!25}$32.35$ & \cellcolor{yellow!25}$26.97$ &  \\
&Plenoxels~\cite{fridovich2022plenoxels} (CVPR'22)   && $26.29 $ &|& \cellcolor{orange!25}$25.46$ & \cellcolor{yellow!25}$27.83$ & $31.09$ & $27.58$ & \cellcolor{orange!25}$21.41$ & \cellcolor{yellow!25}$20.24$ & $30.22$ & $26.48$ &  \\

&Mip-NeRF 360~\cite{barron2022mip} (CVPR'22)  && $-$ &|& $24.59$ & $27.56$ & \cellcolor{yellow!25}$31.34$ & \cellcolor{red!25}$\bm{28.51}$ & $19.84$ & $19.51 $ & \cellcolor{red!25}$\bm{33.49}$ & \cellcolor{red!25}$\bm{27.86}$ & \\

&L2G-NeRF~\cite{chen2023local} (CVPR'23)   && $24.54$ &|& $24.57$ & $24.90$ & $29.27 $ & $23.12$ & $19.02 $ & $19.71$ & $32.25$ & $23.49$ & \\\cline{2-13} 

&BioNeRF (Ours)   && \cellcolor{red!25}$\bm{27.01}$ &|& \cellcolor{red!25}$\bm{26.51}$ & \cellcolor{orange!25}$27.89$ & \cellcolor{red!25}$\bm{32.34}$ & \cellcolor{yellow!25}$27.99$ & \cellcolor{red!25}$\bm{22.23}$ & \cellcolor{red!25}$\bm{20.80}$ & $30.75$ & \cellcolor{orange!25}$27.56$ & \\
\hhline{|-|-|-|-|-|-|-|-|-|-|-|-|-|-|}
\end{tabular}}

\vspace{0.1cm}

\resizebox{.8\textwidth}{!}{
\begin{tabular}{|clcccccccccccc|}
\hhline{|-|-|-|-|-|-|-|-|-|-|-|-|-|-|}
&&&\multicolumn{10}{c}{\textbf{SSIM }$\uparrow$ }&\\
&{\textbf{Method}} && Avg. &|&Fern&Flower&Fortress&Horns &Leaves& Orchids& Room &   T-Rex &\\ \cline{2-13} 
&NeRF~\cite{mildenhall2021nerf} (ACM 21)   && $0.811 $ &|& $0.792$ & $0.827$ & $0.881$ & $0.828$ & $0.690$ & $0.641$ & $0.948$ & $0.880$ &  \\
&TensoRF~\cite{chen2022tensorf} (ECCV'22)   && $0.839$ &|& $0.814$ & \cellcolor{orange!25}$0.871$ & \cellcolor{yellow!25}$0.897 $ & \cellcolor{yellow!25}$0.877$ & \cellcolor{yellow!25}$0.752 $ & $0.649$ & \cellcolor{yellow!25}$0.952$ & \cellcolor{yellow!25}$0.900$ &  \\
&Plenoxels~\cite{fridovich2022plenoxels} (CVPR'22)   && $0.839$ &|& \cellcolor{orange!25}$0.832$ & $0.862$ & $0.885 $ & $0.857$ & \cellcolor{orange!25}$0.760$ & \cellcolor{orange!25}$0.687$ & $0.937 $ & $0.890$ &  \\

&Mip-NeRF 360~\cite{barron2022mip} (CVPR'22)  && $-$ &|& \cellcolor{yellow!25}$0.820$ & \cellcolor{yellow!25}$0.867$ & \cellcolor{orange!25}$0.900$ & \cellcolor{red!25}$\bm{0.909}$ & $0.721 $ & \cellcolor{yellow!25}$0.660$ & \cellcolor{red!25}$\bm{0.965} $ & \cellcolor{red!25}$\bm{0.927}$ & \\

&L2G-NeRF~\cite{chen2023local} (CVPR'23)   && $0.750$ &|& $0.750$ & $0.740$ & $0.840$ & $0.740$ & $0.560 $ & $0.610$ & $0.950$ & $0.800$ & \\\cline{2-13} 

&BioNeRF (Ours)   && \cellcolor{red!25}$\bm{0.861}$ &|& \cellcolor{red!25}$\bm{0.837}$ & \cellcolor{red!25}$\bm{0.873}$ & \cellcolor{red!25}$\bm{0.914}$ & \cellcolor{orange!25}$0.882$ & \cellcolor{red!25}$\bm{0.796}$ & \cellcolor{red!25}$\bm{0.714}$ & \cellcolor{orange!25}$0.956$ & \cellcolor{orange!25}$0.911$ & \\
\hhline{|-|-|-|-|-|-|-|-|-|-|-|-|-|-|}
\end{tabular}}

\vspace{0.1cm}

\resizebox{.8\textwidth}{!}{
\begin{tabular}{|clcccccccccccc|}
\hhline{|-|-|-|-|-|-|-|-|-|-|-|-|-|-|}
&&&\multicolumn{10}{c}{\textbf{LPIPS }$\downarrow$ }&\\
&{\textbf{Method}} && Avg. &|&Fern&Flower&Fortress&Horns &Leaves& Orchids& Room &   T-Rex &\\ \cline{2-13} 
&NeRF~\cite{mildenhall2021nerf} (ACM 21)   && $0.250 $ &|& $0.280$ & $0.219$ & $0.171$ & $0.268$ & $0.316$ & $0.321$ & $0.178$ & $0.249$ &  \\
&TensoRF~\cite{chen2022tensorf} (ECCV'22)   && \cellcolor{orange!25}$0.124$ &|& \cellcolor{orange!25}$0.155$ & \cellcolor{orange!25}$0.106$ & \cellcolor{orange!25}$0.075$ & \cellcolor{orange!25}$0.123$ & \cellcolor{orange!25}$0.153$ & \cellcolor{orange!25}$0.201$ & \cellcolor{yellow!25}$0.082$ & \cellcolor{orange!25}$0.099$ &  \\
&Plenoxels~\cite{fridovich2022plenoxels} (CVPR'22)   && $0.210 $ &|& $0.224$ & $0.179$ & $0.180$ & $0.231$ & \cellcolor{yellow!25}$0.198 $ & \cellcolor{yellow!25}$0.242 $ & $0.192 $ & $0.238$ &  \\

&Mip-NeRF 360~\cite{barron2022mip} (CVPR'22)  && $-$ &|& \cellcolor{yellow!25}$0.210$ & \cellcolor{yellow!25}$0.140$ & $0.117$ & \cellcolor{yellow!25}$0.124$ & $0.231 $ & $0.246$ & $0.115 $ & $0.158$ & \\

&L2G-NeRF~\cite{chen2023local} (CVPR'23)   && \cellcolor{yellow!25}$0.200$ &|& $0.260$ & $0.170$ & \cellcolor{yellow!25}$0.110$ & $0.260$ & $0.330$ & $0.250$ & \cellcolor{orange!25}$0.080$ & \cellcolor{yellow!25}$0.160$ & \\\cline{2-13} 

&BioNeRF (Ours)   && \cellcolor{red!25}$\bm{0.068}$ &|& \cellcolor{red!25}$\bm{0.093}$ & \cellcolor{red!25}$\bm{0.055}$ & \cellcolor{red!25}$\bm{0.025}$ & \cellcolor{red!25}$\bm{0.070}$ & \cellcolor{red!25}$\bm{0.103}$ & \cellcolor{red!25}$\bm{0.122}$ & \cellcolor{red!25}$\bm{0.029}$ & \cellcolor{red!25}$\bm{0.044}$ & \\
\hhline{|-|-|-|-|-|-|-|-|-|-|-|-|-|-|}
\end{tabular}}
\label{t.results_real}
\end{center}
\end{table*}

\subsection{Qualitative evaluation.}
\label{ss.qualitativeEvaluation}

\begin{figure*}[!htb]

    \begin{minipage}{0.25\textwidth}%
		\begin{subfigure}{\linewidth}%
		\centering
		    \includegraphics[width=\textwidth]{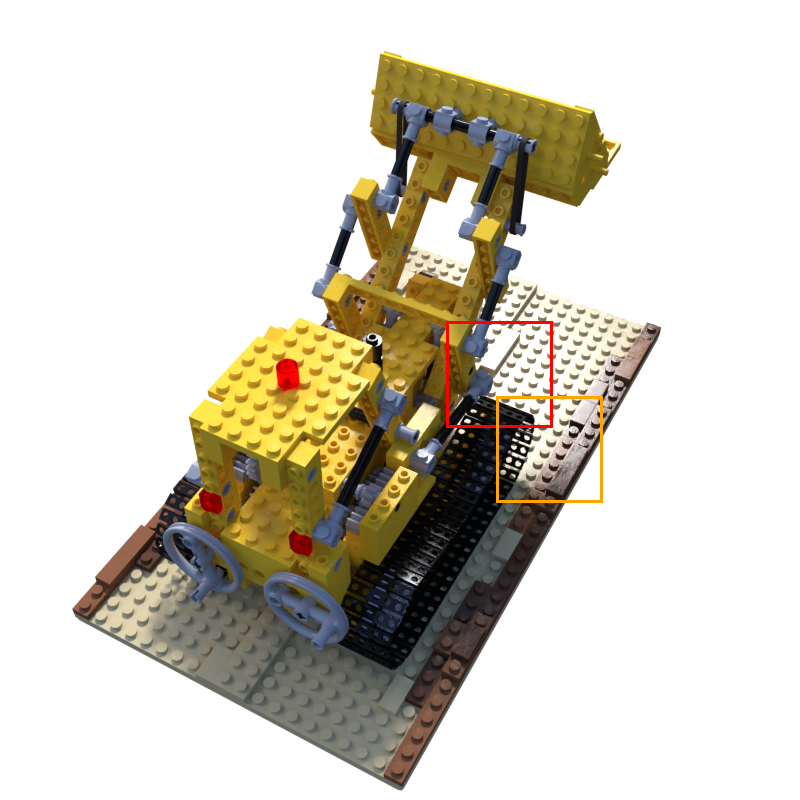}\\
		    Lego
		\end{subfigure}
    \end{minipage}%
    \hspace{0.5mm}
    \begin{minipage}{0.75\textwidth}%
        \begin{minipage}{0.19\textwidth}%
            \begin{subfigure}{\linewidth}%
                \includegraphics[width=\linewidth]{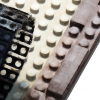}
            \end{subfigure}
        \end{minipage}
        \begin{minipage}{0.19\textwidth}%
            \begin{subfigure}{\linewidth}%
                \includegraphics[width=\linewidth]{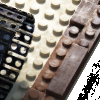}
            \end{subfigure}
        \end{minipage}
        \begin{minipage}{0.19\textwidth}%
            \begin{subfigure}{\linewidth}%
                \includegraphics[width=\linewidth]{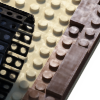}
            \end{subfigure}
        \end{minipage}
        \begin{minipage}{0.19\textwidth}%
            \begin{subfigure}{\linewidth}%
                \includegraphics[width=\linewidth]{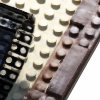}
            \end{subfigure}
        \end{minipage}
        \begin{minipage}{0.19\textwidth}%
            \begin{subfigure}{\linewidth}%
                \includegraphics[width=\linewidth]{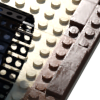}
            \end{subfigure}
        \end{minipage}
        
        \begin{minipage}{0.19\textwidth}%
            \begin{subfigure}{\linewidth}%
            \centering
                \includegraphics[width=\linewidth]{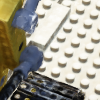}\\
                NeRF
            \end{subfigure}
        \end{minipage}
        \begin{minipage}{0.19\textwidth}%
            \begin{subfigure}{\linewidth}%
            \centering
                \includegraphics[width=\linewidth]{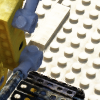}\\
                MipNeRF 360
            \end{subfigure}
        \end{minipage}
        \begin{minipage}{0.19\textwidth}%
            \begin{subfigure}{\linewidth}%
            \centering
                \includegraphics[width=\linewidth]{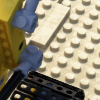}\\
                TensoRF
            \end{subfigure}
        \end{minipage}
        \begin{minipage}{0.19\textwidth}%
            \begin{subfigure}{\linewidth}%
            \centering
                \includegraphics[width=\linewidth]{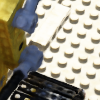}\\
                BioNeRF
            \end{subfigure}
        \end{minipage}
        \begin{minipage}{0.19\textwidth}%
            \begin{subfigure}{\linewidth}%
            \centering
                \includegraphics[width=\linewidth]{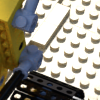}\\
                GT
            \end{subfigure}
        \end{minipage}
    \end{minipage}
    \caption{Qualitative results of BioNeRF and comparison methods, namely NeRF~\cite{mildenhall2021nerf}, Mip-NeRF 360~\cite{barron2022mip},  and TensoRF~\cite{chen2022tensorf}, as well as the ground truth images (GT) on two Blender dataset's scenes.}
    \label{f.comparison_synthetic}
\end{figure*}

This section evaluates the quality of the views generated by the proposed method in terms of visual aspects. Figure~\ref{f.comparison_synthetic} compares BioNeRF against three baselines, i.e., NeRF~\cite{mildenhall2021nerf}, Mip-NeRF 360~\cite{barron2022mip}, and TensoRF~\cite{chen2022tensorf}\footnote{Baseline views obtained from \url{https://nerfbaselines.github.io/}.} considering a Blender dataset's Lego view. Compared to NeRF, BioNeRF presents a result closer to the ground truth, especially if the toy's grid and piston reflections are analyzed in the bottom image. Mip-NeRF 360 shows some black artifacts in the white area on the top image. Regarding TensoRF, BioNeRF shows itself more capable of extracting the reflexible components and reproducing the floor color with better precision.

%Figure~\ref{f.comparison_synthetic} depicts some examples from BioNeRF view synthesis.

% \begin{figure*}[!htb]
%   \centerline{
%     \begin{tabular}{cccc}
%       \includegraphics[width=.25\textwidth]{figs/experiments_synthetic/drums_gt.png} &

%       \includegraphics[width=.25\textwidth]{figs/experiments_synthetic/hotdog_gt.png} &
%       \includegraphics[width=.25\textwidth]{figs/experiments_synthetic/materials_gt.png} &
%       \includegraphics[width=.25\textwidth]{figs/experiments_synthetic/ship_gt.png} \\
%     \end{tabular}}
%   \centerline{
%     \begin{tabular}{cccc}
%       \includegraphics[width=.25\textwidth]{figs/experiments_synthetic/drums.png} &
%       \includegraphics[width=.25\textwidth]{figs/experiments_synthetic/hotdog.png} &

%       \includegraphics[width=.25\textwidth]{figs/experiments_synthetic/materials.png} &

%       \includegraphics[width=.25\textwidth]{figs/experiments_synthetic/ship.png} \\
%       drums & hotdog & materials & ship
%     \end{tabular}}
%   \caption{Ground-truth (top) and synthetic view (bottom) images generated by BioNeRF regarding four Blender dataset's scenes.}
%   \label{f.comparison_synthetic}
% \end{figure*}

\begin{figure*}[!htb]
    \begin{minipage}{0.3\textwidth}%
		\begin{subfigure}{\linewidth}%
		\centering
		    \includegraphics[width=\textwidth]{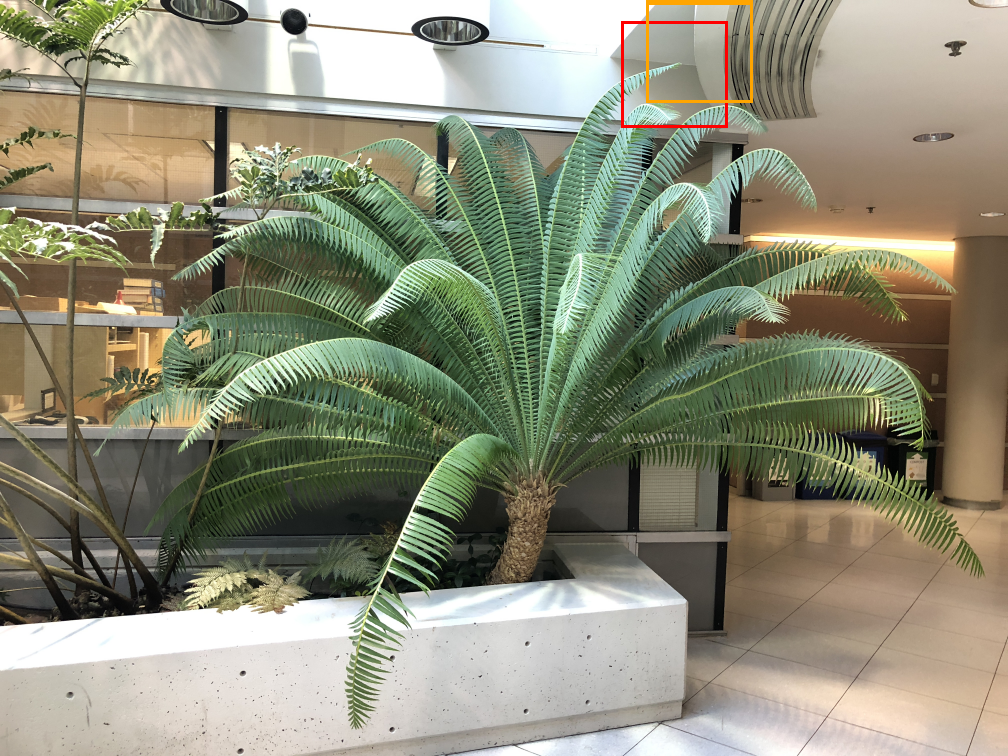}\\
		    Fern
		\end{subfigure}
    \end{minipage}%
    \hspace{0.5mm}
    \begin{minipage}{0.7\textwidth}%
        \begin{minipage}{0.24\textwidth}%
            \begin{subfigure}{\linewidth}%
                \includegraphics[width=\linewidth]{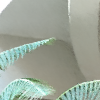}
            \end{subfigure}
        \end{minipage}
        \begin{minipage}{0.24\textwidth}%
            \begin{subfigure}{\linewidth}%
                \includegraphics[width=\linewidth]{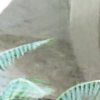}
            \end{subfigure}
        \end{minipage}
        \begin{minipage}{0.24\textwidth}%
            \begin{subfigure}{\linewidth}%
                \includegraphics[width=\linewidth]{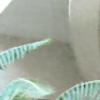}
            \end{subfigure}
        \end{minipage}
        \begin{minipage}{0.24\textwidth}%
            \begin{subfigure}{\linewidth}%
                \includegraphics[width=\linewidth]{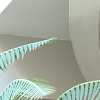}
            \end{subfigure}
        \end{minipage}

        \begin{minipage}{0.24\textwidth}%
            \begin{subfigure}{\linewidth}%
            \centering
                \includegraphics[width=\linewidth]{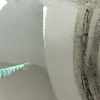}\\
                MipNeRF 360
            \end{subfigure}
        \end{minipage}
        \begin{minipage}{0.24\textwidth}%
            \begin{subfigure}{\linewidth}%
            \centering
                \includegraphics[width=\linewidth]{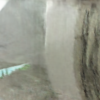}\\
                TensoRF
            \end{subfigure}
        \end{minipage}
        \begin{minipage}{0.24\textwidth}%
            \begin{subfigure}{\linewidth}%
            \centering
                \includegraphics[width=\linewidth]{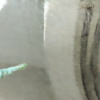}\\
                BioNeRF
            \end{subfigure}
        \end{minipage}
        \begin{minipage}{0.24\textwidth}%
            \begin{subfigure}{\linewidth}%
            \centering
                \includegraphics[width=\linewidth]{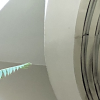}\\
                GT
            \end{subfigure}
        \end{minipage}
    \end{minipage}
    \caption{Qualitative results of BioNeRF and comparison methods, namely, Mip-NeRF 360~\cite{barron2022mip} and TensoRF~\cite{chen2022tensorf}, as well as the ground truth images (GT) on LLFF dataset's Fern scenes.}
    \label{f.comparison_llff}
\end{figure*}

Regarding the LLFF dataset, Figure~\ref{f.comparison_llff} compares BioNeRF against Mip-NeRF 360~\cite{barron2022mip} and TensoRF~\cite{chen2022tensorf} considering a Fern view. In This context, BioNeRF views look cleaner than the instances generated by TensoRF, whose images present some high-frequency artifacts. Regarding Mip-NeRF 360, although the images' resolution looks a bit better than BioNeRF, they show some distorted details considering the air grid on the bottom picture.

\subsection{Ablation study}
\label{ss.ablation}

This section provides an evaluation of the BioNeRF effectiveness over the Blender dataset considering three implemented versions, based on the standard NeRF~\cite{mildenhall2021nerf}, NeRFacto\footnote{\url{https://docs.nerf.studio/nerfology/methods/nerfacto.html}}, and TensoRF~\cite{chen2022tensorf}, which will be referred here as BioNeRF, BioNeRFacto, and BioTensoRF, respectively. The biologically plausible module is implemented in a substructure of NeRF's pipeline called field. In this context, BioNeRF changes are implemented in the Coarse and Fine blocks of NeRF's pipeline\footnote{\url{https://docs.nerf.studio/nerfology/methods/nerf.html}}. Considering BioNeRFacto and BioTensoRF, such changes are implemented in NeRFacto's nerfacto field and TensoRF's field, respectively. 

Table~\ref{t.ablation} provides the average results achieved after training the models for $50k$ iteration. BioNeRF and BioTensoRF outperformed BioNeRFacto results, considering all the evaluation metrics. Considering both methods, BioNeRF outperformed BioTensoRF by $0.001$ considering the SSIM and a far advantage considering the LPIPS metric, While BioTensoRF obtained a slightly better PSNR result. Running the training during more iterations for some dataset scenes, we could observe that BioNeRF sometimes surpasses BioTensoRF PSNR values.

\begin{table}[!htb]
\caption{Average results obtained considering three distinct implementations of BioNeRF over Blender datasets. }
\begin{center}
\renewcommand{\arraystretch}{1.5}
\setlength{\tabcolsep}{6pt}
\resizebox{0.8\columnwidth}{!}{
\begin{tabular}{|clccccc|}
\hhline{|-|-|-|-|-|-|-|}
&{\textbf{Method}} && \textbf{PSNR }$\uparrow$ & \textbf{SSIM }$\uparrow$ & \textbf{LPIPS }$\downarrow$ & \\ \cline{2-6} 
&BioNeRFacto  && $19.01$ & $0.792$ & $0.216$  &\\
&BioTensoRF  &&  $29.15$ & $0.928$ & $0.063$  &\\
&BioNeRF   && $28.81$ & $0.929$ & $0.031$  &\\
\hhline{|-|-|-|-|-|-|-|}
\end{tabular}}
\label{t.ablation}
\end{center}
\end{table}

\section{Discussions, Conclusions, and Future Works}
\label{s.conclusions}

\noindent\textbf{Discussions.} The BioNeRF mechanism reinforces learning through contextual insights to steer the information flow and extract coherent features. Such context is distilled from a memory state updated iteratively in a process that aims to mimic the biological behavior of forgetting by filtering irrelevant information and learning by introducing novel and more relevant discoveries. 

Experiments conducted over two benchmarking datasets for view synthesis, one comprising synthetic and the other real scenes, demonstrate that such a mechanism provided BioNeRF with an outstanding scene representing power, capable of outperforming state-the-art architectures. Regarding such results, BioNeRF achieved the most accuracy overall considering the real scenes dataset over three evaluation metrics for image reconstruction, i.e., PSNR, SSIM, and LPIPS. Regarding the synthetic dataset, BioNeRF obtained the lowest values of LPIPS, outperforming state-of-the-art approaches in a quality measure that best matches human perceptual judgments, thus representing views with better visual quality.

Regarding the model limitations, the primary concern regards the increased number of parameters due to the memory updating system. The problem imposed a more significant challenge due to the limited memory available on the Tesla T4 GPU, i.e., $8$GB. In this context, the experiments were conducted using a batch of directional rays with half the size of the value employed by NeRF to overcome the issue. Such a solution emphasizes the model's robustness since it could obtain state-of-the-art results, even considering such a reduction in instances per iteration.\newline

\noindent\textbf{Conclusions.} The paper introduces BioNeRF for view synthesis. This method implements a mechanism inspired by recent discoveries in neuroscience regarding the behavior of pyramidal cells to model memory and a contextual-based information flow. Such a mechanism allowed BioNeRF to outperform state-of-the-art results over two datasets, considering three metrics for image reconstruction and producing images that best match human perceptual judgments.\newline

\noindent\textbf{Future Works.} Regarding future works, we aim to improve the model by introducing more elements from a biologically plausible perspective, like spiking neurons, implemented in SpikingNeRF~\cite{yao2023spiking}. Additionally, we strive to extend our model to infer in real time and learn from a few shots.
{
    \small
    \bibliographystyle{ieeenat_fullname}
    \bibliography{main}
}


\end{document}